\documentclass{article}

\usepackage{arxiv}

\usepackage[utf8]{inputenc} 
\usepackage[T1]{fontenc}    
\usepackage{hyperref}       
\usepackage{url}            
\usepackage{amsfonts}       
\usepackage{nicefrac}       
\usepackage{microtype}      
\usepackage{graphicx}
\usepackage{wrapfig}

\usepackage{subfig}

\title{Model Development Process}

\author{
  Przemyslaw Biecek\\
  Warsaw University of Technology \& University of Warsaw\\
  \texttt{przemyslaw.biecek@gmail.com}
}

\begin{document}
\maketitle

\begin{abstract}
Predictive modeling has an increasing number of applications in various fields. High demand for predictive models drives creation of tools that automate and support work of data scientist on the model development. To better understand what can be automated we need first a description of the  model life-cycle.

In this paper we propose a generic Model Development Process (MDP). This process is inspired by Rational Unified Process (RUP) which was designed for software development. There are other approached to process description, like CRISP DM or ASUM DM, in this paper we discuss similarities and differences between these methodologies. 

We believe that the proposed open standard for model development will facilitate creation of tools for automation of model training, testing and maintaining.
\end{abstract}


\section{Introduction}

\begin{wrapfigure}{r}{7.0cm}
  \includegraphics[width=7.0cm]{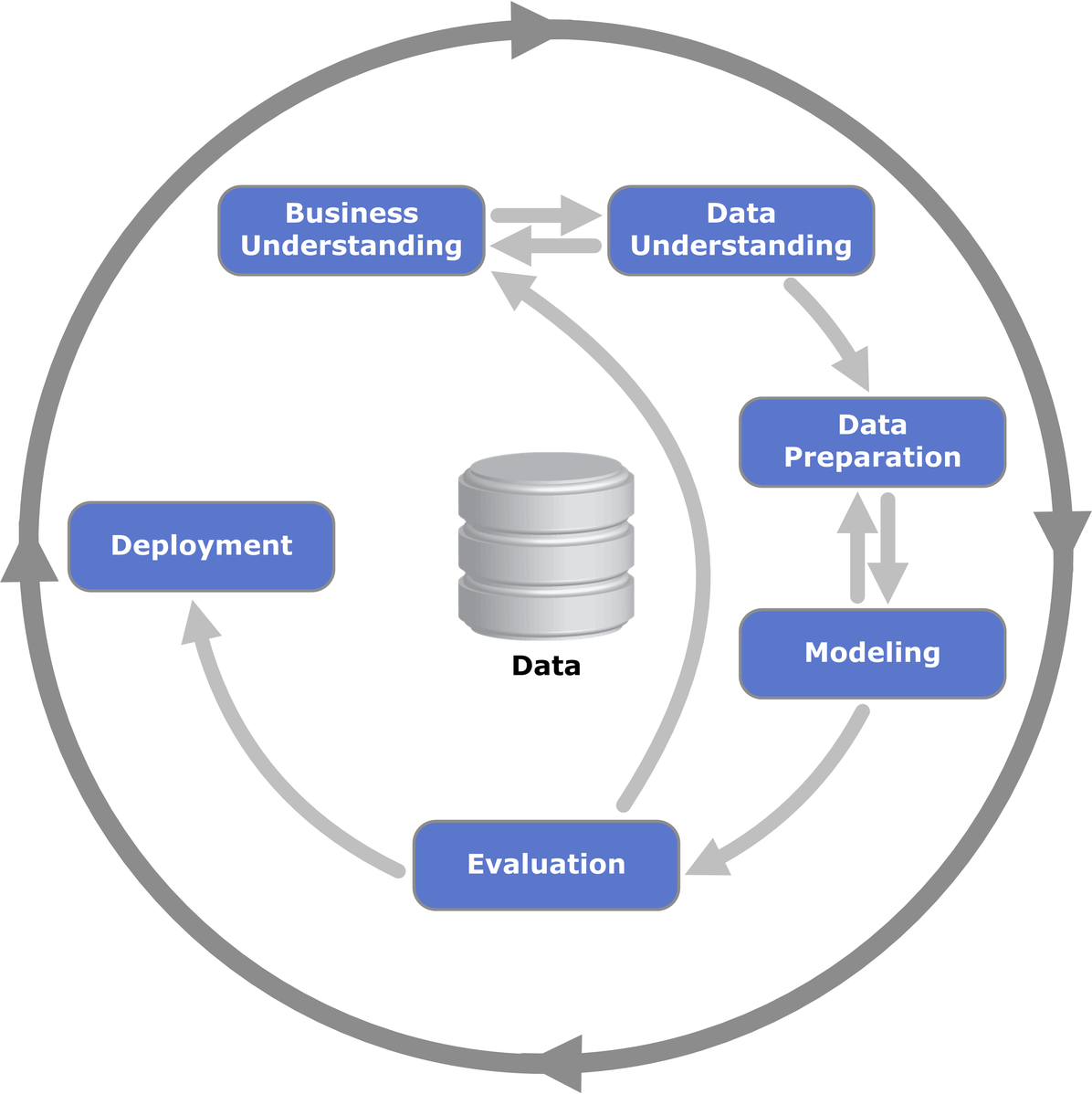}
  \caption{Diagram for \textit{CRISP DM: Cross-industry standard process for data mining} \url{https://en.wikipedia.org/wiki/Cross-industry_standard_process_for_data_mining}.}
  \label{fig:diagramCRISPDM}
\end{wrapfigure}

High demand for predictive models drives creation of tools that automate and support model development, like H2O \cite{h2o}, DataRobot \cite{datarobot}, mljar \cite{mljar}, mlr \cite{mlr}, tpot \cite{tpot}, tidymodels \cite{tidymodels}, scikitlearn \cite{scikitlearn} or others. To better understand the offering of such tools we need better description of the process of model development itself.

One of the most known approach to standardization of data mining projects is the CRISP DM methodology (Cross-industry standard process for data mining) \cite{crisp10} presented in Figure \ref{fig:diagramCRISPDM}. 
CRISP DM was conceived by five companies: Integral Solutions Ltd (ISL), Teradata, Daimler AG, NCR Corporation and OHRA, an insurance company \cite{crispwiki}.

The key component of this approach is the break down of the whole process into six phases: business understanding, data understanding, data preparation, modeling, evaluation and deployment. In general these phases are performed in a consecutive manner, but is it allowed to return to previous phases. The process is iterative as the experience from previous phases helps in consecutive iterations.

\clearpage

\begin{wrapfigure}{r}{7.0cm}
  \includegraphics[width=7.0cm]{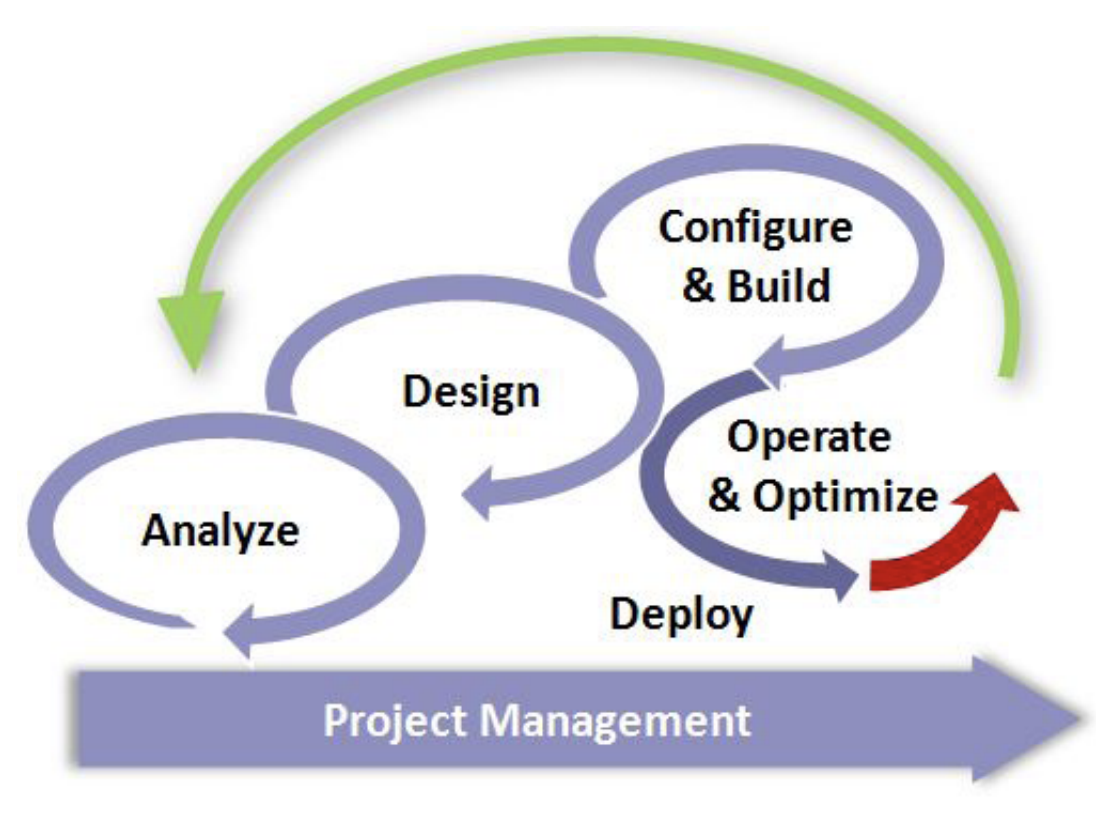}
  \caption{Diagram for \textit{ASUM-DM: Analytics Solutions
Unified Method} \url{ftp://ftp.software.ibm.com/software/data/sw-library/services/ASUM.pdf}.}
  \label{fig:diagramASUM}
\end{wrapfigure}

CRISP DM is a tool agnostic procedure. One of its founders, The ISL company, was later acquired by SPSS which was later acquired by IBM. IBM created a new methodology based on CRISP DM called Analytics Solutions Unified Method for Data Mining/Predictive Analytics (ASUM-DM see Figure \ref{fig:diagramASUM}). Other companies have their own methodologies, for example SAS has SEMMA.

It should be noted that these methods have very wide range of applications, which covers almost any data driven problem, like segmentation or rule based decision systems. In next sections we will focus more on methodologies oriented on development of predictive models.

\begin{figure}[h!]
  \centering
  \subfloat[Diagram from \textit{R for Data Science}, Garrett Grolemund, Hadley Wickham, \url{https://r4ds.had.co.nz/}]{\includegraphics[width=0.48\textwidth]{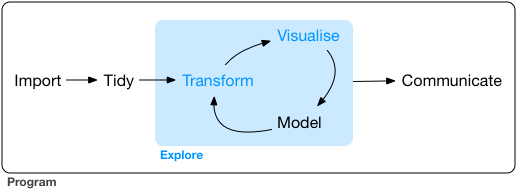}\label{fig:diagramDataScience}}
  \hfill
  \subfloat[Diagram from \textit{A gentle introduction to tidymodels} \url{https://rviews.rstudio.com/2019/06/19/a-gentle-intro-to-tidymodels/}.]{\includegraphics[width=0.48\textwidth]{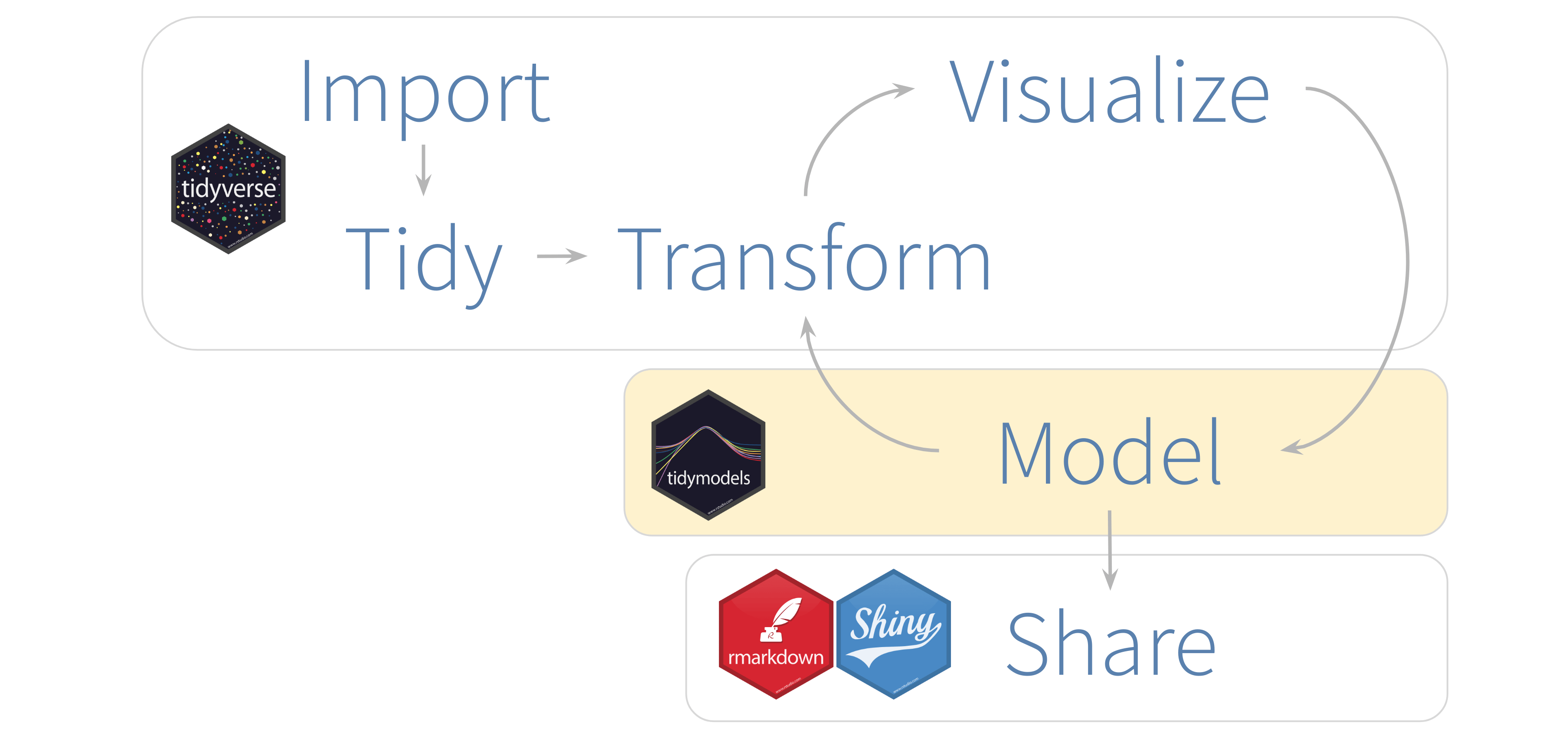}\label{fig:diagramTidyModels}}
  \caption{Diagrams for model development from Tidyverse.}
\end{figure}

CRISP and ASUM are process oriented methodologies. Other approaches cover also methodologies that are more programming oriented. Probably one of the most popular is the approach proposed by Garrett Grolemund and Hadley Wickham in the \textit{R for Data Science}, summarized in Figures \ref{fig:diagramDataScience} and \ref{fig:diagramTidyModels}. This proposition is closely linked with the set of tools developed by RStudio in the tidyverse. 

This approach highlights the iterative nature of data exploration with repetitive steps: Visualise, Model and Transform.
The simplicity of this approach is tempting, yet one needs to remember that in parallel to programming effort we need to think also about other factors like documentation, validation, diagnostics and others.

\begin{wrapfigure}{r}{8.0cm}
  \includegraphics[width=8.0cm]{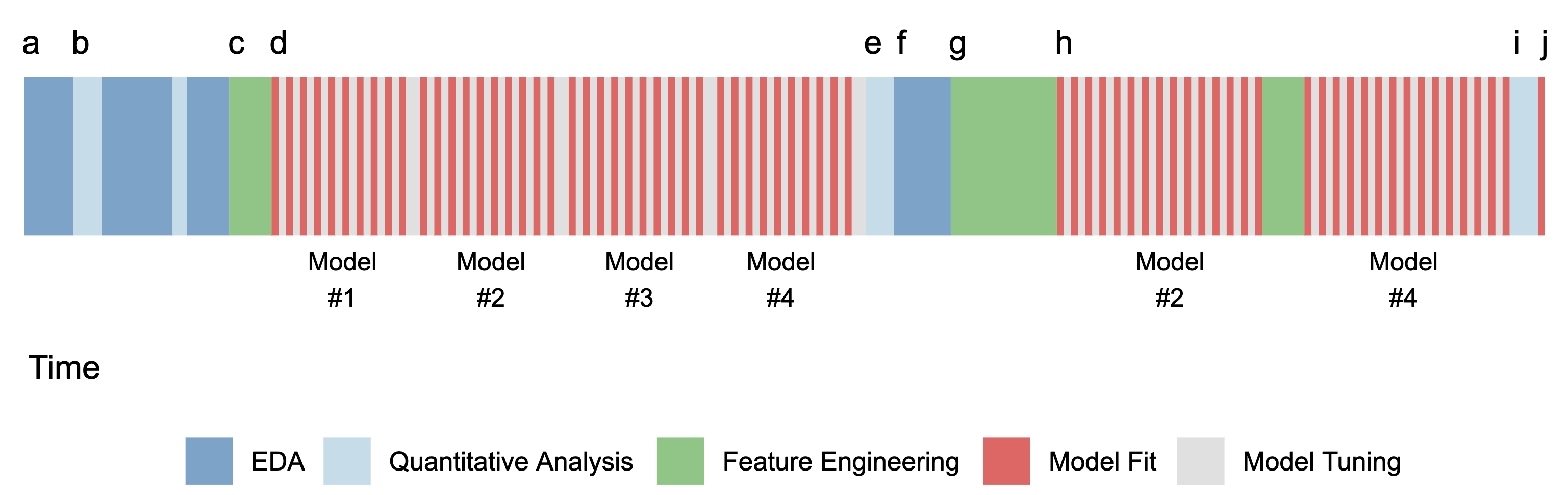}
  \caption{Diagram from \textit{Modeling process according to FES} \url{https://bookdown.org/max/FES/}.}
  \label{fig:diagramFES}
\end{wrapfigure}

Figure \ref{fig:diagramFES} comes from the book \textit{Feature Engineering and Selection} by Max Kuhn and Kjell Johnson. It highlights the iterative feedback loop for model fitting and testing.

In all four presented approaches the iterative nature of the process is highlighted. To some degree this is fully justified as the work with data is almost always iterative as we know more and more and we can use the new knowledge in consecutive iterations. What may be confusing if the composition of consecutive iterations, as it's not the \textit{iterate until convergence} procedure with similar steps repeated before some convergence criteria are met. In fact in data driven tasks consecutive steps have different compositions as some decision is being held and some forking paths are being closed.

\clearpage

\begin{figure}[h!]
\centering
  \includegraphics[width=0.8\textwidth]{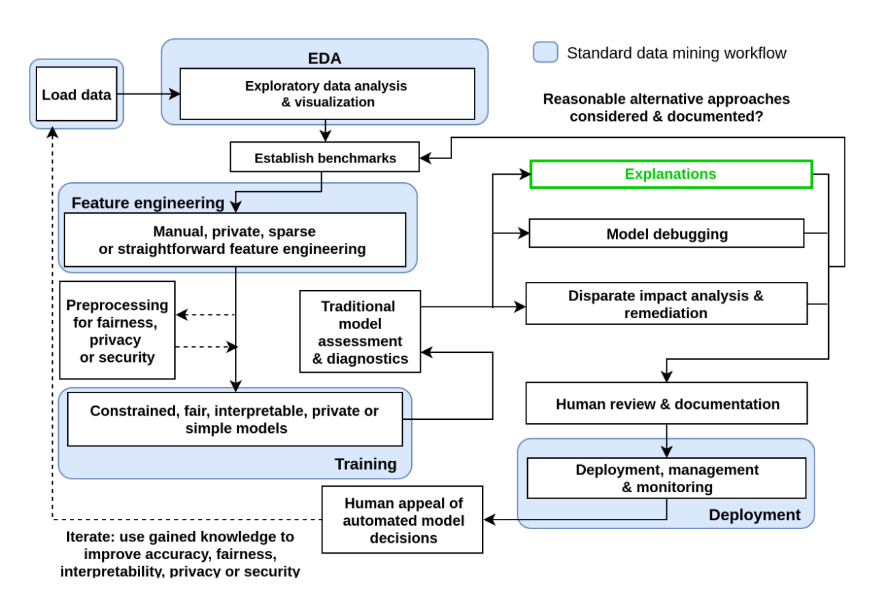}
  \caption{Diagram from \textit{On Explainable Machine Learning Misconceptions and A More Human-Centered Machine Learning}, Patrick Hall,  \url{https://github.com/jphall663/xai_misconceptions/blob/master/xai_misconceptions.pdf}.}
  \label{fig:diagramXAImisconceptions}
\end{figure}

A different approach, more detailed and human centered is presented in the XAI misconceptions paper (see Figure \ref{fig:diagramXAImisconceptions}). There are at least three interesting elements in this diagram. The process is designed explicitly for development of predictive models. Points that require human supervision are directly highlighted (like human review, debugging, explanations). The process is iterative but it is explicit what drives next iterations, e.g. improvements in accuracy, fairness or transparency.

\begin{figure}[h!]
  \centering
  \subfloat[Diagram for \textit{Rational Unified Process} \url{https://en.wikipedia.org/wiki/RUP_hump}]{\includegraphics[width=0.48\textwidth]{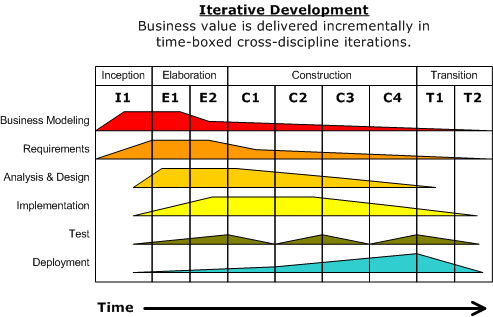}\label{fig:diagramRUP}}
  \hfill
  \subfloat[Diagram for \textit{Spiral model (Boehm, 1988)} \url{https://en.wikipedia.org/wiki/Spiral_model}]{\includegraphics[width=0.48\textwidth]{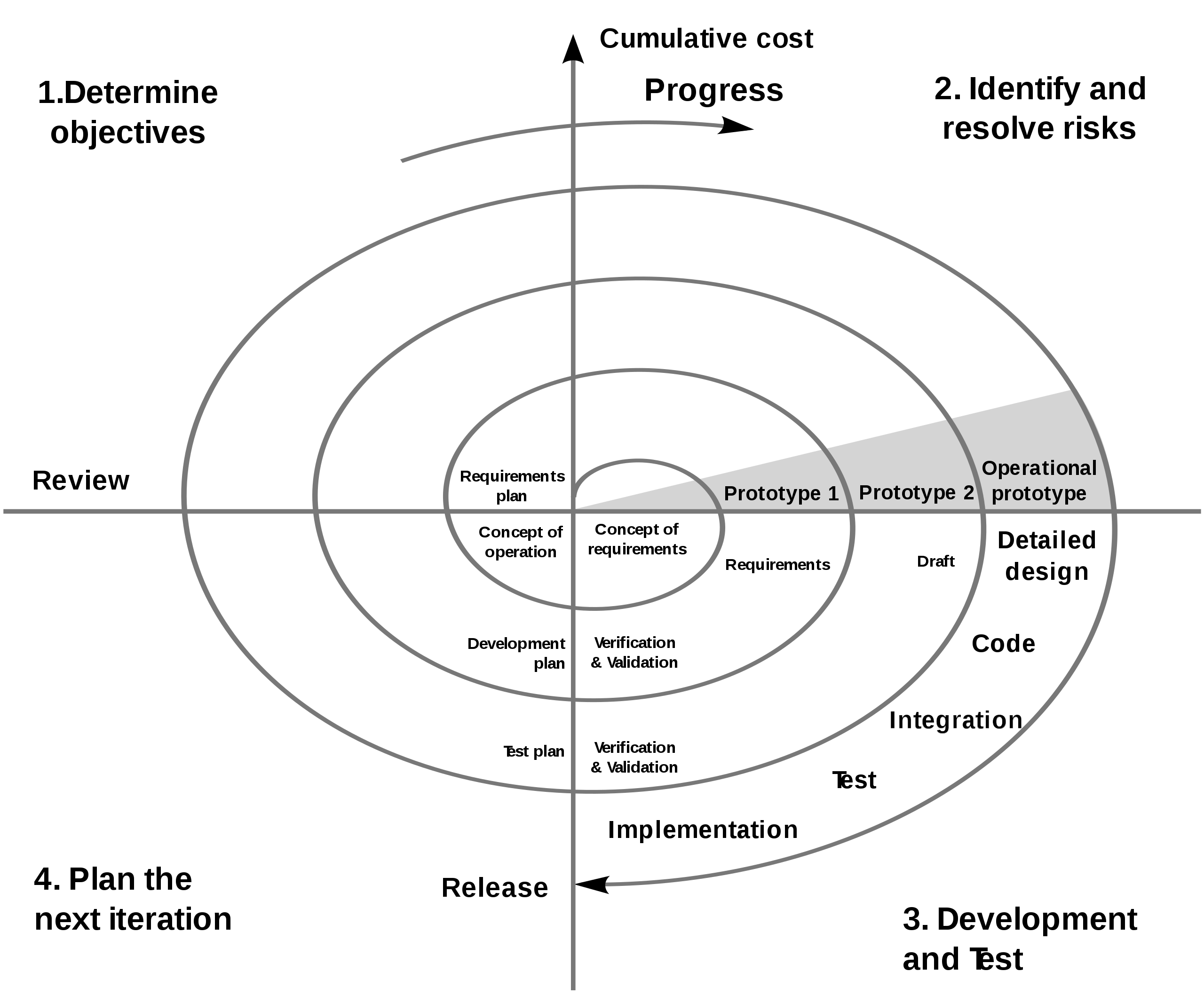}\label{fig:diagramSpiral}}
  \caption{Methodologies for software development .}
\end{figure}

It is interesting to compare these approaches with similar yet more mature field of software development. There are many methodologies that are preferred by different companies depending on scale and nature of the process. Two interesting solutions are the Rational Unified Process \cite{rup1, usdp1} (see Figure \ref{fig:diagramRUP}) developed by the Rational company and the Spiral model \cite{spiral} (see Figure \ref{fig:diagramSpiral}). They replaced the Waterfall model as the iterative nature with quick updated and early checks help to limit number of problems with changing requirements or changes in architecture.

What is interesting in both these proposition is that the process is iterative, but clearly consecutive iterations build on top of previous one. To make it possible, each iteration must be supplemented with some documentation that track the progress of the whole process.

\section{Model Development Process (MDP) in details}

MDP process serves as a skeleton for the \url{http://DrWhy.AI} \cite{DALEX} set of tools. 
This proposition is based on Rational Unified Process and is adapted for development of predictive models.

The described structure is generic. Specific projects may require some changes and adaptations. 

\begin{figure}[h!]
  \includegraphics[width=\textwidth]{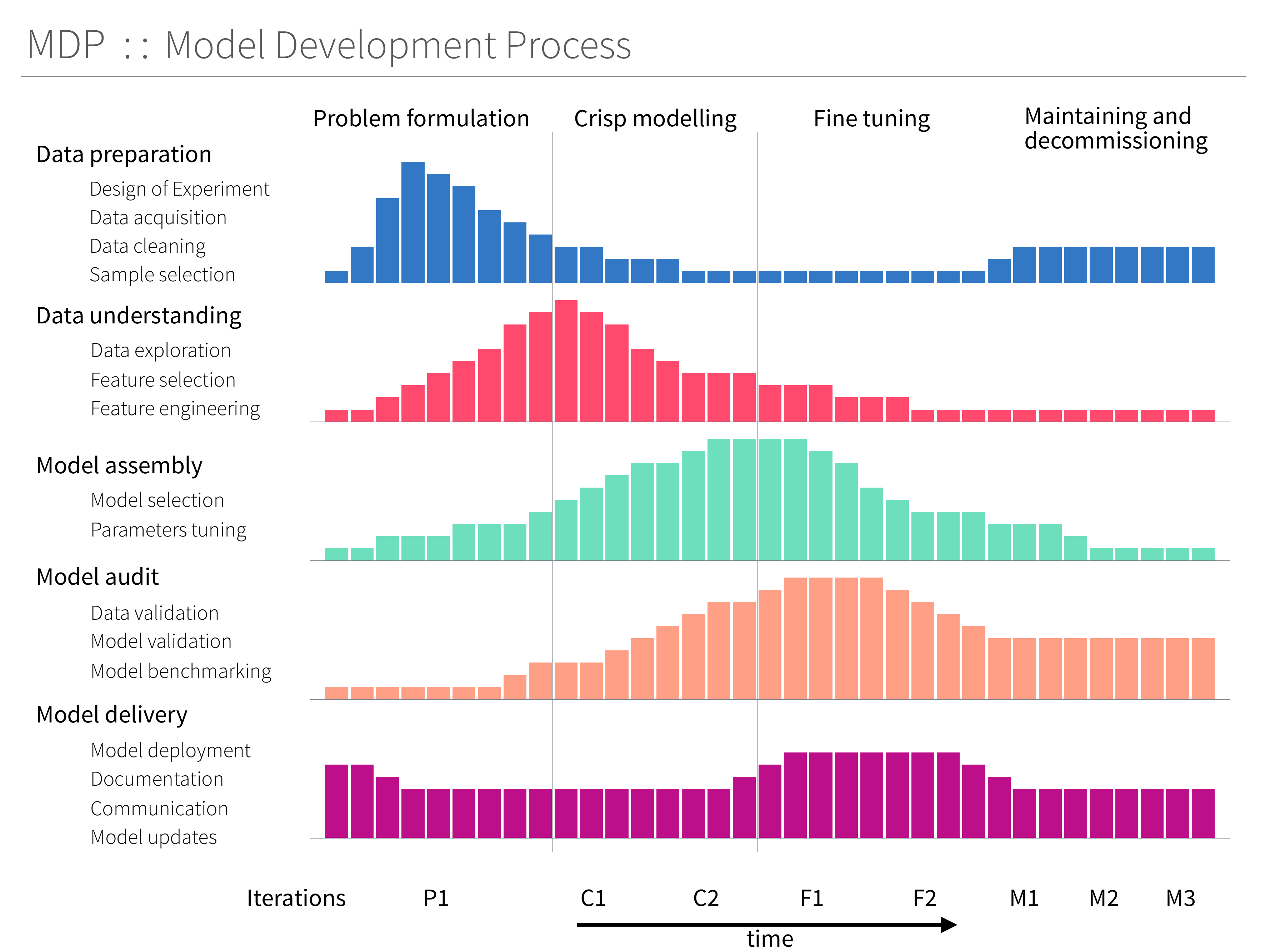}
  \caption{General structure of the \textit{Model Development Process}. Columns correspond to the consecutive phases of model construction. Each phase builds on the knowledge gathered from previous phases. Rows correspond to specific activities that are to be performed in each phase. Heights of bars correspond to the importance of specific activity in each phase.}
  \label{fig:diagramMDP}
\end{figure}

The process is divided into four phases of model life cycle. These phases are used to trace the progress in the model development. As in other methodologies MDP is based on series of iterations. Each phase may be composed out of one or more iterations. 

Tasks that are to be performed in each iteration are listed in rows. As the knowledge about the problem increases every iteration, different tasks require more attention. 

\subsection{Phases and iterations}

Process of model development is divided into four phases of model life cycle from the conception, assembly, tuning till the production.  

\begin{itemize}
    \item Problem formulation --- {What we want to improve? How we measure the improvement?}. The goal of this phase is to precisely define needs for the predictive models, write down \textit{definition of done} and define data set that will be used for training and validation. After this phase we know which performance measures will be used for the assessment of the final model. 
    \item Crisp modeling --- {Generate a prototype quickly. Make sure that this is what client really wants.}. The goal of this phase is to validate the \textit{definition of done}. Here we create first versions of models in order to better understand how close we are to the desired solution. 
    \item Fine tuning --- {Improve the prototype to maximize the desired target measure/minimize loss function.}. The goal of this phase is to tune models identified in the previous phase. Usually in this phase we create large collection of models in order to select the best one (according to some metrics).
    \item Maintaining and decommissioning --- {How model works in the production environment? How it will be updated and how to monitor it's accuracy and fairness. Prepare for model decommissioning one the performance drops too much.}. Developed model go for production. The goal of this phase is to monitor the model and make sure that model performance have not degenerated. Every model will be outdated some day, prepare for the end of model life cycle.
\end{itemize}

In more complex projects one phase may be divided into set of iterations. The maintaining phase usually is composed out of series of periodic health-checks.

\subsection{Tasks}

Phases corresponds to the general progress in model development, while tasks corresponds to programming and analytic activities that needs to be done in each iteration. Importance of particular tasks is changing along model life-cycle. In the diagram we showed some general patterns, but specific problems may require more effort in some of these phases.

\begin{itemize}
    \item Data preparation --- { Activities needed for selection of the training, test and validation data. }
    \begin{itemize}
        \item Data acquisition. Sometimes data needs to be read from file, from database or from some stream. Sometimes we need to scrap data from website. Sometimes one datset is not enough and we need to acquire more (maybe paid) datasets that will be combined.
        \item Data cleaning. Different data sources have different quality. Sometimes some values needs recoding, errors in the data needs to be spotted. 
        \item Sample selection. Good model requires good and carefully selected dataset. Outliers needs to be handled. If data is not balanced or is heterogeneous this needs to be handled, typically through oversampling, undersampling or segmentation.
    \end{itemize}
    \item Data understanding --- {Activities needed for getting some lever of familiarity with the data, needed for further modeling.}   
\begin{itemize}
        \item Data exploration. What are uni- and multi- variate distributions. What are relation between dependent variable and explanatory variables. Do we have missing values. How strong is the correlation between different features. 
        \item Feature selection. Which variables shall be included in the model. Assessment of their predictive power independently and in groups of other variables.
        \item Feature engineering. How variables should be encoded. Factors may need some recoding, continuous variables may need some transformations of discretisation. Groups of variables may need blending.
    \end{itemize}
    \item Model assembly --- { Activities needed for the construction of the model. }
    \begin{itemize}
        \item Model selection. There is an increasing number of different procedures for model construction. Further, new models can be created as a combination of other models. 
        \item Parameters tuning. Most procedures for model constructions are parametrized. Different strategies may be employed to identify best set of parameters.
    \end{itemize}
    \item Model audit --- {Activities needed for monitoring model performance, fairness and stability.}
    \begin{itemize}
        \item Data validation. Is there a change in the structure of the data, distributions of variables or relation structure?
        \item Model validation. Is there a change in model performance between training, test and validation data. Is there a change in performance in the new batch of validation data? Is there any issue in model fairness?
        \item Model benchmarking. How good is a given model in comparison to other models?
    \end{itemize}
    \item Model delivery --- {Activities needed for model release. }
    \begin{itemize}
        \item Model deployment. Model needs to be put in the production environment keeping same version of dependent libraries.
        \item Documentation. Decisions that lead to the final model needs to be saved. Model and data used for training should be clearly defined.  Documentation shall be gathered and expanded through the full model lifetime.
        \item Communication. Reports, charts, tables, all artifact that are used to consult the model  with the client in a easy to understand way.
        \item Model updates. With new batches of data one may plan model retraining to adjust for recent data. This phase is more common for time-series models.
    \end{itemize}
\end{itemize}

\section{Acknowledgments}

I would like to thank Patrick Hall, Max Kuhn, Wit Jakuczun and Alicja Gosiewska for valuable comments and discussions related to MDP.

\bibliographystyle{apalike}
\bibliography{modelDevelopmentProcess}

\begin{thebibliography}{}

\bibitem[Biecek, 2018]{DALEX}
Biecek, P. (2018).
\newblock {DALEX: Explainers for Complex Predictive Models in R}.
\newblock {\em Journal of Machine Learning Research}, 19(84):1--5.

\bibitem[Bischl et~al., 2016]{mlr}
Bischl, B., Lang, M., Kotthoff, L., Schiffner, J., Richter, J., Studerus, E.,
  Casalicchio, G., and Jones, Z.~M. (2016).
\newblock {mlr}: Machine learning in r.
\newblock {\em Journal of Machine Learning Research}, 17(170):1--5.

\bibitem[Boehm, 1988]{spiral}
Boehm, B. (1988).
\newblock {\em A Spiral Model of Software Development and Enhancement}.

\bibitem[Chapman et~al., 1999]{crisp10}
Chapman, P., Clinton, J., Kerber, R., Khabaza, T., Reinartz, T., Shearer, C.,
  and Wirth, R. (1999).
\newblock {\em The CRISP-DM 1.0 Step-by-step data mining guide}.

\bibitem[DataRobot, 2019]{datarobot}
DataRobot (2019).
\newblock {\em DataRobot: automated machine learning platform}.

\bibitem[H2O, 2019]{h2o}
H2O (2019).
\newblock {\em H2O: in-memory platform for distributed, scalable machine
  learning}.

\bibitem[Jacobson et~al., 1999]{usdp1}
Jacobson, I., Booch, G., and Rumbaugh, J. (1999).
\newblock {\em The Unified Software Development Process}.

\bibitem[Kruchten, 1998]{rup1}
Kruchten, P. (1998).
\newblock {\em The Rational Unified Process: An Introduction}.

\bibitem[Kuhn and Hadley, 2018]{tidymodels}
Kuhn, M. and Hadley, W. (2018).
\newblock {\em tidymodels: Easily Install and Load the 'Tidymodels' Packages}.
\newblock R package version 0.0.2.

\bibitem[Olson et~al., 2016]{tpot}
Olson, R.~S., Urbanowicz, R.~J., Andrews, P.~C., Lavender, N.~A., Kidd, L.~C.,
  and Moore, J.~H. (2016).
\newblock {\em Applications of Evolutionary Computation: 19th European
  Conference, EvoApplications 2016, Porto, Portugal, March 30 -- April 1, 2016,
  Proceedings, Part I}, chapter Automating Biomedical Data Science Through
  Tree-Based Pipeline Optimization, pages 123--137.
\newblock Springer International Publishing.

\bibitem[Pedregosa et~al., 2011]{scikitlearn}
Pedregosa, F., Varoquaux, G., Gramfort, A., Michel, V., Thirion, B., Grisel,
  O., Blondel, M., Prettenhofer, P., Weiss, R., Dubourg, V., Vanderplas, J.,
  Passos, A., Cournapeau, D., Brucher, M., Perrot, M., and Duchesnay, E.
  (2011).
\newblock Scikit-learn: Machine learning in {P}ython.
\newblock {\em Journal of Machine Learning Research}, 12:2825--2830.

\bibitem[Ploski, 2019]{mljar}
Ploski, P. (2019).
\newblock {\em mljar: automated machine learning platform}.

\bibitem[Wikipedia, 2019]{crispwiki}
Wikipedia (2019).
\newblock {\em CRISP DM: Cross-industry standard process for data mining}.

\end{thebibliography}

\end{document}